\documentclass[10pt, a4paper]{article}
\usepackage{lrec}
\usepackage{multibib}
\newcites{languageresource}{Language Resources}
\usepackage{graphicx}
\usepackage{tabularx}
\usepackage{soul}
\usepackage{epstopdf}
\usepackage[utf8]{inputenc}
\usepackage{hyperref}
\usepackage{xstring}
\usepackage{color}
\usepackage[table,xcdraw]{xcolor}

\title{\textbf{Cross-Lingual Machine Speech Chain for \\Javanese, Sundanese, Balinese, and Bataks Speech Recognition and Synthesis}}
\name{Sashi Novitasari$^{1}$, Andros Tjandra$^{1}$, Sakriani Sakti$^{1,2}$, Satoshi Nakamura$^{1,2}$}
\address{$^{1}$Nara Institute of Science and Technology, Japan\\
	     $^{2}$RIKEN Center for Advanced Intelligence Project AIP, Japan\\
	     \{sashi.novitasari.si3, tjandra.ai6, ssakti,s-nakamura\}@is.naist.jp\\}

\abstract{
Even though over seven hundred ethnic languages are spoken in Indonesia, the available technology remains limited that could support communication within indigenous communities as well as with people outside the villages. As a result, indigenous communities still face isolation due to cultural barriers; languages continue to disappear. To accelerate communication, speech-to-speech translation (S2ST) technology is one approach that can overcome language barriers. However, S2ST systems require machine translation (MT), speech recognition (ASR), and synthesis (TTS) that rely heavily on supervised training and a broad set of language resources that can be difficult to collect from ethnic communities. Recently, a machine speech chain mechanism was proposed to enable ASR and TTS to assist each other in semi-supervised learning. The framework was initially implemented only for monolingual languages. In this study, we focus on developing speech recognition and synthesis for these Indonesian ethnic languages: Javanese, Sundanese, Balinese, and Bataks. We first separately train ASR and TTS of standard Indonesian in supervised training. We then develop ASR and TTS of ethnic languages by utilizing Indonesian ASR and TTS in a cross-lingual machine speech chain framework with only text or only speech data removing the need for paired speech-text data of those ethnic languages.
\\
\newline }

\begin{document}

\maketitleabstract

\section{Introduction}
Indonesia, which has some of the world's most diverse religions, languages, and cultures \cite{abas_1987_1,bertrand_2003_1,hoon_2006_1}, consists of approximately 17,500 islands with 300 ethnic groups and 726 native languages \cite{tan_2004_1}. Roughly ten percent of the world's
languages are spoken in Indonesia, making it one of the most multilingual nations in the world. In the amidst of such a large number of local languages, \textit{Bahasa Indonesia}, the national language, functions as a bridge that connects Indonesian people. \textit{Bahasa Indonesia} is a unity language, which was coined by Indonesian nationalists in 1928 and became a symbol of national identity during the struggle for independence
in 1945. Since then, the Indonesian language is increasingly being spoken as a second language by the majority of its population. The decision to choose Indonesian as a unity language is one great success story of national language policy \cite{sneddon_2003_1,paauw_2009_1}.

Worldwide globalization is encouraging people to learn and speak languages that are prominent in global communities. \textit{Bahasa
Indonesia} is now more commonly spoken as a first language. Some younger Indonesians are also speaking English as a second language.
Although using \textit{Bahasa Indonesia} as the unity language is helping them face globalization, multilingualism in Indonesia faces
a catastrophe. The number of speakers of Indonesian ethnic languages is decreasing. It is predicted that Indonesia might
shift from a multilingual nation to a monolingual society, threatening the existence of ethnic languages \cite{cohn_2014_1}.

Among its 726 ethnic languages, only thirteen have more than a million speakers, accounting for about 70\% of the total population in Indonesia.
These ethnic languages include Javanese, Sundanese, Malay, Madurese, Minangkabau, Bataks, Bugisnese, Balinese, Acehnese, Sasak, Makasarese, Lampungese, and Rejang \cite{lauder_2005_1}. The remaining 713 languages have a total population of only 41.4 million speakers, and the majority of these have very small numbers of speakers \cite{riza_2008_1}. For example, 386 languages are spoken by 5,000 or less; 233 have 1,000 speakers or less; 169 languages have 500 speakers or less; and 52 have 100 or less \cite{gordon_2005_1}. These languages are facing various degrees of language endangerment \cite{CRY}. Several attempts have addressed preserving ethnic languages, including national projects on the use
of ethnic languages in schools. Unfortunately, the available technology that could support communication within indigenous communities as well as with people outside the villages is limited. Indigenous communities face the digital divide and isolation due to cultural barriers. Languages continue to be threatened.

Speech-to-speech translation (S2ST) technology \cite{nakamura_2009_1,sakti_2013_1}, which is innovative and essential, enables people to communicate in their native
languages. S2ST recognized the speech of the source language into the text, translate the text to the target language, and synthesizes back to speech waveforms. This overall technology involves research in machine translation (MT), automatic speech recognition (ASR), and text-to-speech synthesis (TTS). However, the advanced development of these technologies relies heavily on supervised training and a broad set of language resources, including speech and corresponding transcriptions of the source and target languages. Unfortunately, the amount of
available Indonesian ethnic language data is limited, and preparing a large amount of paired speech and text data is expensive.

Recently, a machine speech chain framework was proposed for the semi-supervised development of ASR and TTS systems \cite{tjandra_2017_1}. This framework was motivated by the human speech chain mechanism \cite{denes_1993_1}, which is a feedback loop phenomenon between speech production and a hearing system that occurs when humans speak. In fact, humans do not separately learn to speak and listen using supervised training with a large amount of paired data. By simultaneously listening and speaking, they monitor their own volume and articulation and gradually improve their speaking capability, making it consistent with their intentions.

In the machine speech chain, both ASR and TTS components are pre-trained
in supervised training using a limited amount of labeled data. By establishing a feedback closed-loop between the ASR (listening
component) and the TTS (speaking component), both components can assist each other in unsupervised learning. Therefore, they can be
trained without requiring a large number of speech-text paired data. Previous machine speech chain studies \cite{tjandra_2017_1,tjandra_2018_1,tjandra_2019_1}, however, only utilized the framework for monolingual model training and
unsupervised training with a large number of unlabeled or unpaired data. The framework remains unutilized for a cross-lingual task
with a limited number of unpaired data, such as ethnic languages.

In this work, we utilize the machine speech chain framework in a cross-lingual way to construct speech recognition and synthesis for
the following four ethnic Indonesian languages: Javanese, Sundanese, Balinese, and Bataks. Although scant research has addressed the development of ASR for Indonesian ethnic languages, one study developed ASR for those languages using a statistical approach with a hidden Markov model and a Gaussian mixture model (HMM-GMM) \cite{sakti_2014_1}. However, no ASR construction with a sequence-to-sequence deep-learning approach has been made. Furthermore, no previous TTS study exists for Indonesian ethnic languages. The previous study also still requires paired speech and a corresponding transcription of the ethnic languages for supervised adaptation. In contrast, we develop both ASR and TTS of those ethnic languages based on sequence-to-sequence deep-learning architectures. We first separately train ASR and TTS of standard Indonesian in supervised training. We then train the ASR and TTS of those ethnic languages by utilizing Indonesian ASR and TTS in a cross-lingual machine speech chain framework with limited text or speech of those ethnic languages. This choice allows us to construct ASR and TTS for those languages, even without paired data for them.

\section{Overview of Indonesian and Indonesian Ethnic Languages}

Here, we briefly introduce the Indonesian and Indonesian ethnic languages.

\subsection{Indonesian Language}
The Indonesian language, \textit{Bahasa Indonesia}, is derived from the Malay dialect, which was the lingua franca of Southeast Asia \cite{quinn_2001_1}. Bahasa Indonesia is closely related to the Malay spoken in Malaysia, Singapore, Brunei, and some other areas. It is the largest member of the Austronesian language family. The only difference is that Indonesia (a former Dutch colony) adopted the Van Ophuysen orthography in 1901; Malaysia (a former British colony) adopted the Wilkinson orthography in 1904. In 1972, the governments of Indonesia and
Malaysia agreed to standardize ``improved'' spelling, which is now in effect on both sides. Even so, modern Indonesian and modern
Malaysian are as different from one another as are Flemish and Dutch \cite{tan_2004_1}.

Many words in the Indonesian vocabulary reflect the historical influence of the various colonial cultures that occupied and influenced the
archipelago. Indonesian words have borrowed heavily from Indian Sanskrit, Chinese, Arabic, Portuguese, Dutch, and English \cite{jones_2007_1}. Unlike Chinese, it is not a tonal language; it has no declensions or conjugations. It has no changes in nouns or adjectives for different gender, number, or case. Verbs do not take different forms to show number, person, or tense. A time adverb or question word can be placed at either the
front or the end of sentences. Since plural is often expressed by reduplication, Indonesian sentences have reduplication words. It is
also a member of the agglutinative language family, denoting a complex range of prefixes and suffixes that are attached to base
words that can result in very long words \cite{sakti_2004_1}.

The standard Indonesian language is mostly used in such formal written settings as books, newspapers, and television/radio news
broadcasts. Although the earliest records in Malay inscriptions are syllable-based and written in Arabic script, modern Indonesian is
phonetic-based written in Roman script \cite{ALW}. It only uses 26 letters, as in the English/Dutch alphabet.

\subsection{Indonesian Ethnic Languages}
In this study, we chose to work with four ethnic languages: Javanese, Sundanese, Balinese, and Bataks. Since a large number of the population speak them, the data collection of their native speakers remain possible to reach. However, despite their significant speech communities, the primary usage of these languages is gradually being subsumed by \textit{Bahasa Indonesia}. The Javanese, Sundanese, Balinese, and Bataks languages
also suffer from the inadequate intergenerational transmission, since they are often not being passed on to the next generation. It is pointed out that even such languages are at risk of language endangerment \cite{cohn_2014_1}.

\subsubsection{Javanese}

Javanese is a member of Malayo-Polynesian, which is a branch of the Austronesian language family. It is spoken by Javanese people from
the central and eastern parts of Java, which has almost 100 million native speakers \cite{cohn_2014_1} (more than 42\% of Indonesia's
population). It is also spoken in Suriname and New Caledonia to which it was originally brought by Javanese workers who were
transferred from Indonesia by the Dutch. Javanese transcription is called \textit{Aksara Hanacaraka}\footnote{The following is the
official site of Aksara Jawa: http://hanacaraka.fateback.com/}. Aksara means transcription in Indonesia. It consists of 20 basic
scripts or letters called \textit{Carakan}. One \textit{Carakan} stands for a syllable with a consonant and an inherent vowel. To create another sound with other Javanese vowels, an additional script called \textit{Sandhangan} to define the vowel is need.

\subsubsection{Sundanese}
The Sundanese language, which is also a Malayo-Polynesian language spoken by the people who live on the western third of Java island,
has almost 40 million native speakers who represent about 15\% of Indonesia's population \cite{bauer_2007_1}. Modern Sunda transcription is called Aksara Sunda. Similar to Aksara Hanacaraka, Aksara Sunda also has a basic alphabet, vowels, and punctuation that change the phonemes and the
basic punctuation\footnote{http://en.wikipedia.org/wiki/Sundanese\_alphabet}.

\subsubsection{Balinese}
Balinese is the native language of Bali island. It is spoken by more than three million people and is also a member of the
Malayo-Polynesian language family \cite{bauer_2007_1}. The Balinese script\footnote{http://en.wikipedia.org/wiki/Balinese\_alphabet} is undoubtedly derived from the Devanagari and Pallava scripts from India. Its shape resembles southern Indian scripts like Tamil. However, most Balinese people only use the Balinese language for oral communication, often mixing it with Indonesian in their daily speech. In 2011, the Bali Cultural Agency estimates that the number of people still using Balinese does not exceed one million, which is only one-fourth of the total Bali population. Balinese is mostly spoken in social and cultural interactions; Indonesian, however, is increasingly the language of commerce, in schools and public places \cite{horstman_2016_1}.

\subsubsection{Bataks}
The Batak languages are a subgroup of the Austronesian languages spoken by the Batak people in the Indonesian province of North
Sumatra and its surrounding areas \cite{bauer_2007_1}. The Batak tribes are descendants of a powerful Proto-Malayan people who mainly lived in the northern region of Sumatera Island. There are several subtribes and clans in Batak tribes. The Toba subtribe has
the largest population, followed (in no particular order) by the Karo, Simalungun, Pakpak-Dairi, Angkola-Mandailing, and Nias (Niha)
peoples. The Batak tribe has its own writing system, which dates back to the 13th century. The Batak people call their writing
system \textit{Surat Batak} (Surat = letters/writings)\footnote{http://www.ancientscripts.comt}.

\section{Speech and Text Data Resources}
In this study, we utilized several data resources including Indonesian and Indonesian ethnic language corpora. The details are
described below.

\subsection{Indonesian Data Resources}
\label{sec:idn_data}
The Indonesian speech dataset was developed by the R\&D Division of an Indonesian telecommunications company (PT. Telekomunikasi Indonesia) in collaboration with ATR Japan under the Asia Pacific Telecommunity (APT) project \citelanguageresource{sakti_2004_1,sakti_2008_1}. The corpus consists of 80.5 hours of speech spoken by multiple speakers. The following are its details of the data resources.

\subsubsection{Text}
 The transcriptions in the Indonesian corpus were originally constructed from two sources: daily news passages and telephone
services dialogs. The daily news sentences were compiled from the most widely read Indonesian newspapers: \textit{Kompas} and \textit{Tempo}. The text sentences from telephone service dialogs consist of language commonly required for such services as tele-home security, hotel
reservations, billing information, e-Government status tracking, and hearing-impaired telecommunication (HITS) services. From the
two sources, phonetically-balanced sentences were selected based on a greedy search algorithm \cite{zhang_2003_1}, resulting in 5,668 sentences of the total: 3,168 sentences from news passages and 2,500 sentences from telephone service dialogs.

\subsubsection{Speech}
The speech audio, which was recorded using the clean sentences of the above text data, was spoken by 400 speakers (200 males and 200 females) from various regions in Indonesia. 
The recording was conducted in parallel for both clean and telephone speech, recorded at a respective sampling frequency of 16 kHz and 8 kHz. However, in this study, only the speech utterances with 16 kHz sampling frequency were used.
During the recording stage, all the speakers
were instructed to speak in standard Indonesia without any ethnic accents. Each speaker spoke 110 sentences from the news passage,
which resulted in 44,000 speech utterances (43.35 hours of speech), and 100 sentences from the telephone services dialogs, which
resulted in 40,000 utterances (36.15 hours of speech). The size of the speech data was 84,000 utterances: around 80.5 hours of speech.

\subsection{Indonesian Ethnic Data Resources}
\label{sec:el_data}
The Indonesian ethnic languages covered in this work include Javanese, Sundanese, Balinese, and Bataks, all of which were collected and recorded in a previous study \citelanguageresource{sakti_2013_1}.

\subsubsection{Text}
The text data consist of 225 sentences selected from online newspapers and magazines in their ethnic languages: Pejabar-Semangat for Javanese\footnote{www.penjebarsemangat.co.id}, Sunda-News for Sundanese\footnote{sundanews.com}, Bali-Post for Balinese\footnote{www.balipost.co.id}, and Halo-Moantondang for Bataks\footnote{halomoantondang.wordpress.com}. The sentences were
selected by cleaning the raw text and limiting the sentences to those with a graphemically-balanced structure based on a greedy
search algorithm \cite{zhang_2003_1}.

We also translated fifty Indonesian sentences from the ATR basic travel expression corpus (BTEC) \citelanguageresource{kikui_2003_1} dataset into these ethnic languages
by native speakers of the corresponding language. We created 275 sentences for each ethnic language, comprised of 1,100 text transcriptions.

\subsubsection{Speech}
The speech data were recorded from ten native speakers (five males and five females) for each ethnic language. Each speaker spoke 225
graphemically balanced sentences and 100 Indonesian-ethnic language parallel sentences, which consist of 50 original Indonesian
sentences and 50 that were translated into ethnic languages. However, in this study, we only used 275 sentences of
the ethnic languages, and removed the 50 original Indonesian sentences. The speech was recorded at a 48-kHz sampling rate with 16-bit resolution. In the experiment, all the speech utterances were downsampled into 16 kHz. The speech data were composed of 2,750 utterances with ten speakers for each language, comprised of 11,000 speech utterances.

\section{Speech Chain}
\label{sec:spchain}
\subsection{Human Speech Chain}
\begin{figure}[h]
  \centering
  \includegraphics[width=0.48\textwidth]{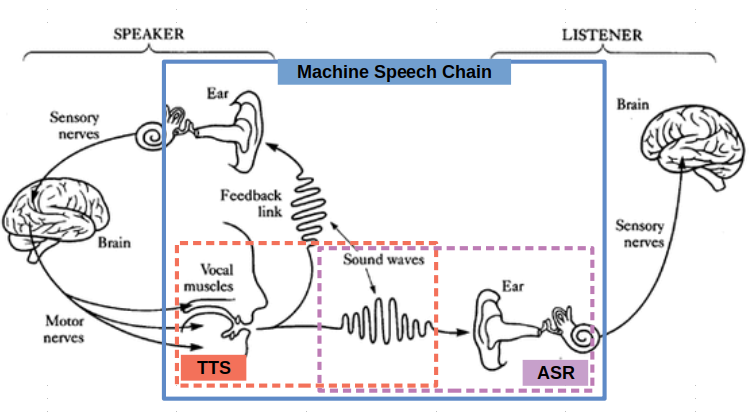}
  \caption{Human speech chain and corresponding components in machine speech chain (Denes et al., 1993).}
  \label{fig:human_spchain}
\end{figure}

\begin{figure}[h]
  \centering
  \includegraphics[width=0.48\textwidth]{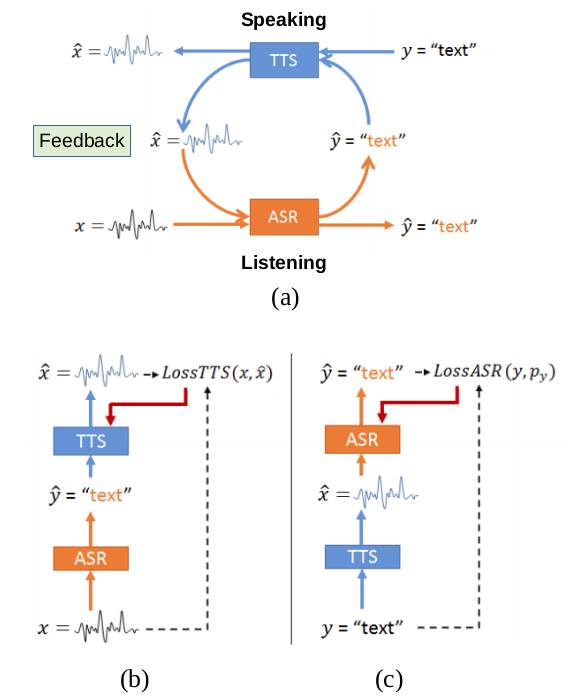}
  \caption{Overview of machine speech chain: (a) Feedback loop connects ASR and TTS based on concept of speaking while listening in human speech chain process. Loop can be unrolled into two processes: (b) from ASR to TTS and (c) from TTS to ASR (Tjandra et al., 2017).}
  \label{fig:machine_spchain}
\end{figure}

The human speech chain was previously introduced \cite{denes_1993_1} as a phenomenon in human communication (Fig.~\ref{fig:human_spchain}). In a conversation, the speaker's utterance is heard by the listener and also the speaker herself. A feedback chain is established among the speaker's hearing system, her brain, and the speech production system. When the speaker listens to her speech, she compares it to
her intended quality and uses it to improve its quality in the next timestep. The speaking and listening processes occur simultaneously
and are continually repeated until the end of the utterance.

\subsection{Machine Speech Chain}
Inspired by the human speech chain, a machine speech chain was proposed to jointly train ASR and TTS models in semi-supervised
learning. An overview of the machine speech chain \cite{tjandra_2017_1} is shown in Fig.~\ref{fig:machine_spchain}(a). The framework consists of a
sequence-to-sequence ASR and a sequence-to-sequence TTS. Sequence-to-sequence networks are deep-learning architecture that include an encoder and a decoder with an attention mechanism. The machine speech chain establishes a loop that connects ASR to TTS and TTS to ASR. The components are trained with a semi-supervised approach that consists of two stages: supervised and unsupervised training.

\subsubsection{Supervised Training}
In the supervised training stage, both ASR and TTS are trained independently using labeled or speech-text paired data. Each model
is trained by minimizing the loss between the predicted output sequence and the ground-truth sequence. Supervised training acts as
knowledge initialization in each component.

\subsubsection{Unsupervised Training}
The unsupervised training stage utilizes models that have already been supervisedly pre-trained and further trained in the speech
chain mechanism using unlabeled data (only speech or text). To learn from the unlabeled data, both ASR and TTS need to support
each other, bypassing feedback through a loop that connects them. The loop consists of two unrolled processes: (1) given only speech
data, the unrolled process performs from ASR to TTS; (2) given only text data, the unrolled process performs from TTS to ASR.

The unrolled process from ASR to TTS is shown in Fig.~\ref{fig:machine_spchain}(b). Given only speech feature
sequences, ASR generates its transcription, and TTS reconstructs the speech based on ASR output. The loss is calculated by comparing
the TTS-generated speech and the original speech. The unrolled process from TTS to ASR is shown in Fig.~\ref{fig:machine_spchain}(c). Here given only text transcription, TTS synthesizes the speech from it, and ASR transcribes the speech from TTS. The loss is calculated by
comparing the transcription from ASR and the original text.

\section{Machine Speech Chain for \\Indonesian Ethnic Languages}
\label{sec:ethchain}
Here we utilized the machine speech chain in a cross-lingual setting. Both the ASR output and the TTS input are represented as character sequences to avoid out-of-vocabulary words, especially during the unsupervised training phase. The ASR in this work does not utilize any language model.
Below are the steps of the training process.

\subsection{Step 1: Supervised training of standard Indonesian ASR and TTS}

\begin{figure}[h]
  \centering
  \includegraphics[width=0.48\textwidth]{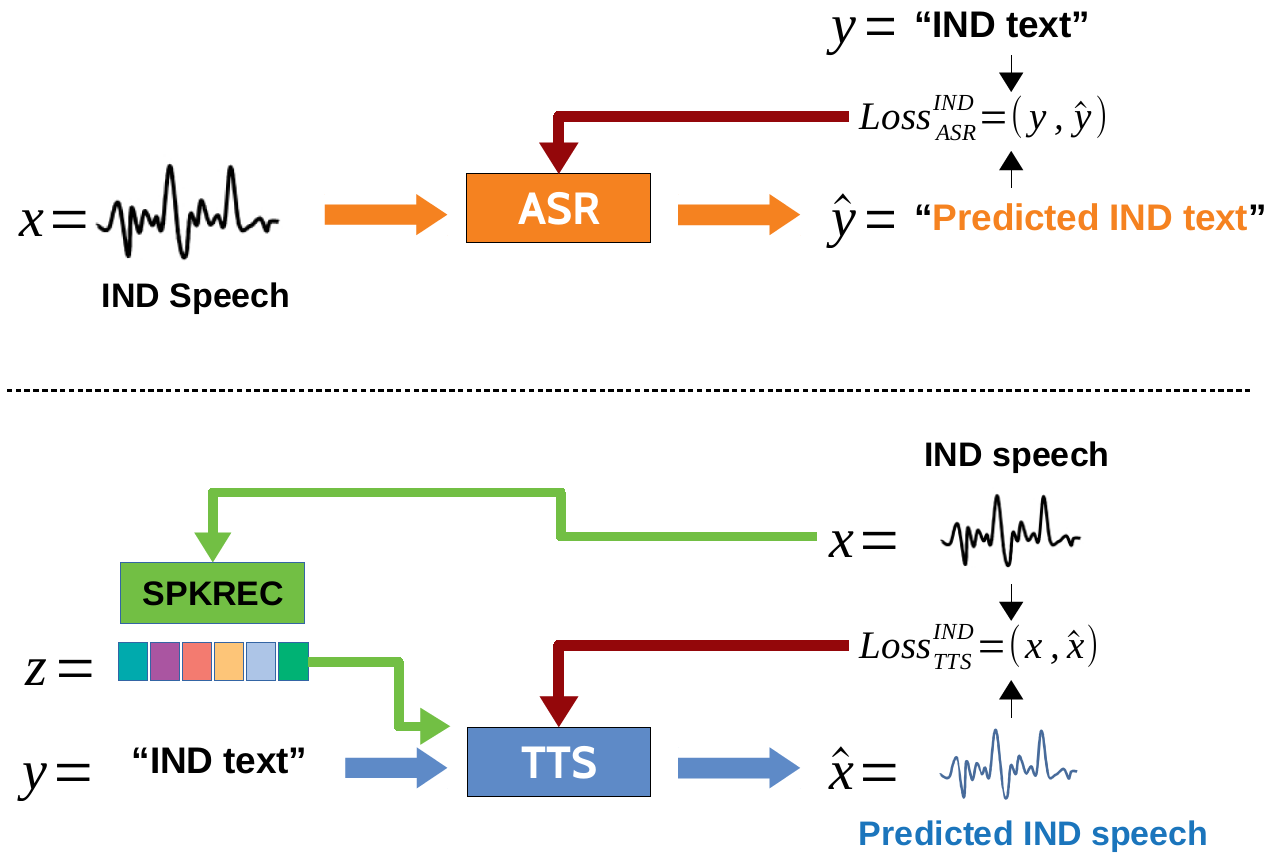}
  \caption{ASR and TTS supervised training using paired speech and text of Indonesian data. Both models were trained separately.}
  \label{fig:idn_supervised}
\end{figure}

First, we supervisedly train the standard Indonesian ASR and TTS using Indonesian speech-text paired data (see Section~\ref{sec:idn_data}). Here the Indonesian language serves as prior knowledge for the system. In this stage, since Indonesian speech-text paired data are available, both the Indonesian ASR and TTS components are trained independently, as seen in Fig.~\ref{fig:idn_supervised}. The ASR takes a sequence of Indonesian speech features $\mathbf{x}^{(IND)}$
and learns to transcribe it into text $\mathbf{\hat{y}}^{(IND)}$. The TTS takes Indonesian text sentence $\mathbf{y}^{(IND)}$ and learns to generate speech $\mathbf{\hat{x}}^{(IND)}$. The speech data here consist of multi-speaker speech. Therefore, the TTS input also includes speaker embedding vector $z= SPKREC(\mathbf{x}^{(IND)})$, and so the synthesized speech can be compared to the ground speech with appropriate voice characteristics, as proposed in a previous machine speech chain study \cite{tjandra_2018_1}.

\subsection{Step 2: Unsupervised training of Javanese, Sundanese, Balinese, Bataks ASR and TTS}

\begin{figure}[h]
  \centering
  \includegraphics[width=0.48\textwidth]{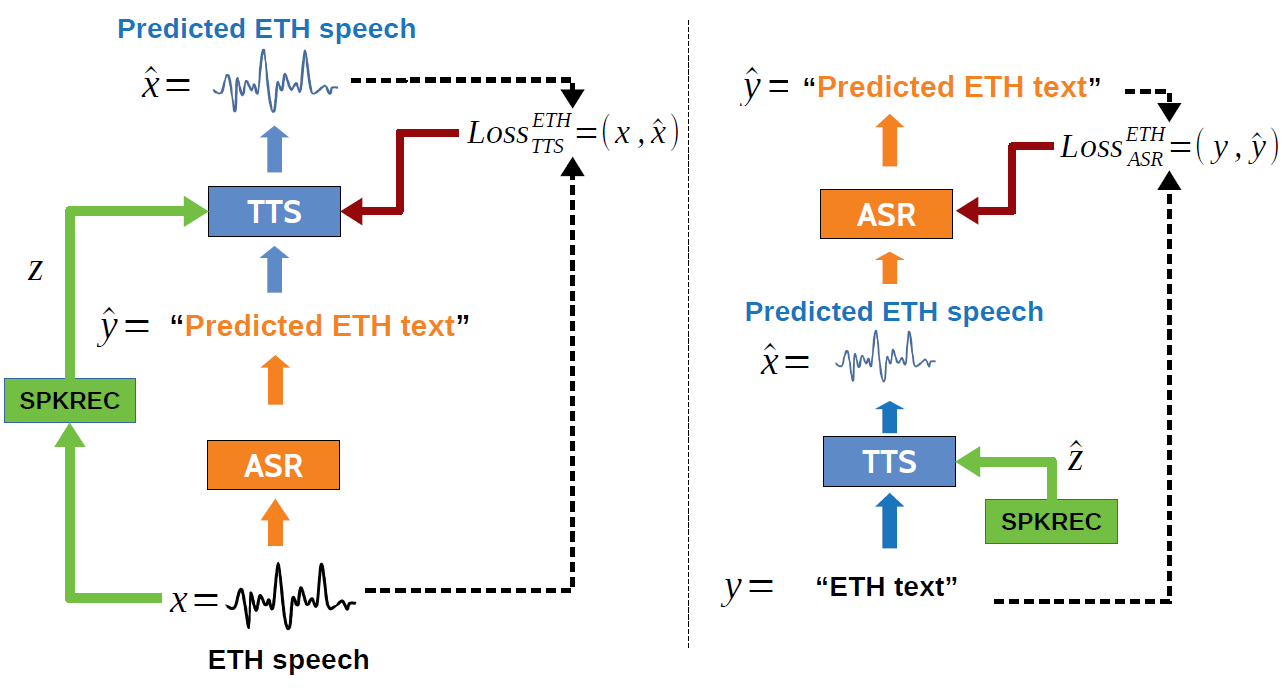}
  \caption{ASR and TTS unsupervised training using unpaired data of Indonesian ethnic languages: text only data (TTS-to-ASR process) and speech only data (ASR-to-TTS process). \textit{ETH} is Indonesian ethnic language.}
  \label{fig:ethnic_unsupervised}
\end{figure}

In this phase, we utilized the previously pre-trained Indonesian ASR and TTS in the speech-chain architecture to unsupervisedly train the ASR and TTS of Javanese, Sundanese, Balinese, and Bataks. Since the available data are minimal, we combined all the unlabeled data (only text or only speech) from the Javanese, Sundanese, Balinese, and Bataks corpus (see Section~\ref{sec:el_data}). This phase is shown in Fig.~\ref{fig:ethnic_unsupervised}, and the following are the details of the unrolled processes.
      \begin{enumerate}
        \item \textbf{Unsupervisedly train the system given only the text transcription of the Indonesian ethnic languages (Closed-loop from TTS to ASR)}:
          In this process, the provided ground information only consists of text data from the four ethnic languages. The TTS attempted to synthesize a sequence of speech features in particular ethnic language $\mathbf{\hat{x}}^{(ETH)}$, based on given text $\mathbf{y}^{(ETH)}$ and the speaker embedding vector. Speaker embedding vector $\hat{z}$ is generated based on speech sampled from the available speech data ($\mathbf{\tilde{x}}$). After the speech synthesis, ASR attempted to transcribe back ethnic language text 
          $\mathbf{\hat{y}}^{(ETH)}$ based on the TTS output $\mathbf{\hat{x}}^{(ETH)}$. The loss is then calculated by comparing the original ethnic text transcription $\mathbf{y}^{(ETH)}$ and the ASR output $\mathbf{\hat{y}}^{(ETH)}$; and perform back-propagation to improve the ASR.

        \item \textbf{Unsupervisedly train the system given only the speech utterances of the Indonesian ethnic languages (Closed-loop from ASR to TTS)}:
        This process uses only speech data from the four ethnic languages. First, ASR took original ethnic speech $\mathbf{x}^{(ETH)}$ and predicted its transcription $\mathbf{\hat{y}}^{(ETH)}$. The TTS then reconstructed ethnic language speech $\mathbf{\hat{x}}^{(ETH)}$ by processing the text generated by ASR and speaker embedding vector $z= SPKREC(\mathbf{x}^{(ETH)})$. The loss is calculated by comparing the original ethnic speech utterances $\mathbf{x}^{(ETH)}$ and the generated TTS speech output $\mathbf{\hat{x}}^{(ETH)}$; and perform back-propagation to improve the TTS.
      \end{enumerate}

\section{Experimental Set-Up}

\subsection{Training, Validation, and Test Datasets}

For supervised training on both the ASR and TTS components of the Indonesian language, we chose 10\% of the speech-text paired data
with 40 speakers for testing. On the remaining speech-text paired data with 360 speakers, 20\% of the data were randomly selected for
validation or development sets, and 80\% of the rest was used as a training set.

For unsupervised training of both the ASR and TTS components of the Javanese, Sundanese, Balinese, and Bataks languages, we randomly
chose 50 unpaired data from four speakers (200 speech utterances or text transcription only) of each language as a test set. On the remaining data of 225 speech or text unpaired data with six speakers (1350 speech utterances or text transcription only), 10\% of the data were randomly selected for validation or development sets, and 90\% of the rest was used for training.

\subsection{Speech Features and Text Representation}

We extracted two different sets of speech features. First, we applied pre-emphasis (0.97) on the raw waveform and then extracted
the log-linear spectrogram with a 50-ms window, 12.5-ms steps, and a 2048-point short-time Fourier transform (STFT) with the Librosa
package \cite{librosa}. Second, we extracted the log Mel-spectrogram with an 80 Mel-scale filterbank. For our TTS model,
we used both log-linear and log-Mel spectrogram for the first and second outputs. For our ASR and speaker recognition components, we
used the log-Mel spectrogram for the encoder input.

The text utterances were tokenized as Indonesian characters and mapped into a 33-character set: 26 alphabetic letters (a-z), three
punctuation marks ('.-), and four special tags, $\langle$\texttt{noise}$\rangle$, $\langle$\texttt{spc}$\rangle$,
$\langle$\texttt{s}$\rangle$, and $\langle$\texttt{/s}$\rangle$ as noise, space, start-of-sequence, and end-of-sequence tokens,
respectively. Both the ASR input and the TTS output shared the same text representation in the training and inference stages.

\subsection{ASR and TTS Systems}

\subsubsection{ASR Component}

For the ASR system, we used a standard sequence-to-sequence model with an attention module. On the encoder sides, the input
Mel-spectrogram features were projected by a fully connected layer with a 512 hidden units and LReLU function. Later, the results were
processed by three bidirectional LSTMs (Bi-LSTM) with 256 hidden units for each LSTM (512 hidden units for Bi-LSTM). To reduce the
memory consumption and processing time, we used hierarchical sub-sampling \cite{graves2012supervised,bahdanau2016end} on all
three Bi-LSTM layers and reduced the sequence length by a factor of 8. On the decoder sides, we projected one-hot encoding from the
previous character into a 256-dims continuous vector with an embedding matrix, followed by a unidirectional LSTM with 512 hidden
units. For the attention module, we used standard content-based attention \cite{bahdanau2014neural}. In the decoding phase, the
transcription was generated by beam-search decoding (size=5), and we normalized the log-likelihood score by dividing it by its own
length to prevent the decoder from favoring shorter transcriptions.

\subsubsection{TTS Component}

For the TTS system, we followed the previously proposed TTS architecture \cite{tjandra_2018_1}, which is a modification from
TTS Tacotron \cite{wang2017tacotron}. The hyperparameters for the basic structure were generally identical as those of the original
Tacotron, except ReLU is replaced with the LReLU ($\alpha=0.01$) function. For the CBHG module, we used $K=8$ filterbanks instead of
16 to reduce the GPU memory consumption. For the decoder sides, we deployed two LSTMs instead of a GRU with 256 hidden units. For each
time-step, our model generated four consecutive frames to reduce the number of steps in the decoding process.

\subsubsection{Speaker Recognition Component}

For the speaker recognition system, we used the DeepSpeaker model \cite{li2017deep} and followed the original hyper-parameters in
that previous paper. However, since data are often scarce in the Indonesian and Indonesian ethnic languages, we utilized a
DeepSpeaker, which was already pre-trained on the Wall Street Journal CSR Corpus of English language (SI84 set with 83 unique speakers) \citelanguageresource{paul1992design}. Thus, the model
was expected to generalize effectively across all of the remaining unseen speakers to assist the TTS and speech chain training. We
used the Adam optimizer with a learning rate of $5e-4$ for the ASR and TTS models and $1e-3$ for the DeepSpeaker model. All of our
models in this paper were implemented with PyTorch \cite{paszke2017automatic}.

\subsubsection{Systems Evaluation Metrics}
We evaluated the ASR system performance based on the character error rate (CER) of the output.
The CER calculation follows the Eq.~\ref{eqn:asr_loss}.

 \begin{equation}
	CER= \frac{S+D+I}{N} \times 100\%
 \label{eqn:asr_loss}
 \end{equation}

$S$, $D$, and $I$ denote the numbers of character substitutions, deletions, and insertions respectively, and $N$ denotes the number of characters in the reference text.
It is similar to the calculation of word error rate (WER), with a difference that WER is calculated based on word tokens, while CER is based on character tokens.

For the TTS system, we evaluated its performance by calculating the L2 norm-squared on log-Mel spectrogram of reference speech (${\mathbf{x}}$) and TTS speech ($\mathbf{\hat{x}}$) using Eq.~\ref{eqn:tts_loss}:

 \begin{equation}
 Loss_{TTS} = \frac{1}{T} \sum_{t=1}^{T}(x_t-\hat{x}_t)^2
 \label{eqn:tts_loss}
 \end{equation}
 where $T$ is the length of speech.

\section{Experimental Results}
\begin{table*}[t]
\footnotesize
\begin{center}
\begin{tabular}{|l|l|c|c|c|c|c|}
\hline
\multicolumn{2}{|c|}{\textbf{Training}} & \multicolumn{5}{c|}{\textbf{Testing}}\\
\hline
\multicolumn{1}{|c|}{\textbf{ASR System}} & \multicolumn{1}{c|}{\textbf{Data}} & \multicolumn{1}{c|}{\textbf{Javanese}} & \multicolumn{1}{c|}{\textbf{Sundanese}} & \multicolumn{1}{c|}{\textbf{Balinese}} & \multicolumn{1}{c|}{\textbf{Bataks}}& \multicolumn{1}{c|}{\textbf{Avr}}\\
    \hline
	\hline
Baseline IND & Sup IND (Sp+Txt) & 107.26 & 90.70 & 97.98 & 109.85 & 101.45\\
 \hline
\hline
Proposed1 IND+ETH & Sup IND (Sp+Txt) + Unsup ETH (Txt Only) & 63.73 & 63.04 & 70.80 & 72.79 & 67.59\\
Proposed2 IND+ETH & Sup IND (Sp+Txt) + Unsup ETH (Sp+Txt) & 31.96 & 31.97 & 27.00 & 37.37 & 32.08\\
 \hline
	\hline
Topline IND+ETH  & Sup IND (Sp+Txt) + Sup ETH (Sp+Txt) & 20.20 & 17.89 & 15.41 & 26.69 & 20.05\\
	\hline 
\end{tabular}
\caption{ASR performances by character error rate (CER (\%)). Here, Indonesian language is denoted as \textit{IND}, while Indonesian ethnic language is denoted as \textit{ETH}. \textit{Sup} is supervised learning and \textit{Unsup} is unsupervised learning.}
\label{tb:ASRResults}
\end{center}
\end{table*}

\begin{table*}[t]
\footnotesize
\begin{center}
\begin{tabular}{|l|l|c|c|c|c|c|}
\hline
\multicolumn{2}{|c|}{\textbf{Training}} & \multicolumn{5}{c|}{\textbf{Testing}}\\
\hline
\multicolumn{1}{|c|}{\textbf{TTS System}} & \multicolumn{1}{c|}{\textbf{Data}} & \multicolumn{1}{c|}{\textbf{Javanese}} & \multicolumn{1}{c|}{\textbf{Sundanese}} & \multicolumn{1}{c|}{\textbf{Balinese}} & \multicolumn{1}{c|}{\textbf{Bataks}}& \multicolumn{1}{c|}{\textbf{Avr}}\\
    \hline
	\hline
Baseline IND & Sup IND (Sp+Txt) & 1.016 & 1.247 & 1.129 & 1.254 & 1.162\\
 \hline
\hline
Proposed IND+ETH & Sup IND (Sp+Txt) + Unsup ETH (Sp+Txt) & 0.547 & 0.531 & 0.560 & 0.510 & 0.537\\
 \hline
	\hline
Topline IND+ETH  & Sup IND (Sp+Txt) + Sup ETH (Sp+Txt) & 0.415 & 0.470 & 0.478 & 0.399 & 0.441\\
	\hline 
\end{tabular}
\caption{TTS performances in L2 norm-squared on log-Mel spectrogram. Here, Indonesian language is denoted as \textit{IND}, while Indonesian ethnic language is denoted as \textit{ETH}. \textit{Sup} is supervised learning and \textit{Unsup} is unsupervised learning.}
\label{tb:TTSResults}
\end{center}
\end{table*}

We separately evaluated the ASR and TTS using test sets of four ethnic languages: Javanese, Sundanese, Balinese, and Bataks.

\subsection{ASR Evaluation}


Table~\ref{tb:ASRResults} shows the character error rate (CER (\%)) performance of the ASR systems from multiple scenarios evaluated on those ethnic language test data. 
In the first block, we supervisedly trained our baseline system just using the speech-text paired data of the Indonesian language. 
Unfortunately, the number of errors (refer to $S+D+I$ in Eq. 1) in the recognition output exceeds the number of characters in the reference ($N$), resulting in a CER that above 100\%.
This indicates that directly using the Indonesian ASR is difficult for recognizing ethnic languages.
In the second block, we showed our proposed approach that utilized cross-lingual speech-chain framework (see the training process in Section~\ref{sec:ethchain}). We utilized the previously pre-trained Indonesian ASR and TTS. We then develop ASR of those ethnic languages given only text or both text and speech (but unpaired). The results reveal that given only text data, the proposed system improved the performance and reduced the average CER from 101.45\% to 67.59\%, which is 33.86\% absolute reduction. If both text and speech data exist (but
unpaired), we might further reduce the average CER to 32.08\%, which is 69.37\% absolute reduction from the baseline. In the last
block, we showed the topline system in which the system was trained using both paired speech and the text of the Indonesian and Indonesian ethnic
languages in a supervised manner. The system's performance achieved an average CER of 20.05\%.

\subsection{TTS Evaluation}

Similar to the ASR evaluation, Table~\ref{tb:TTSResults} shows the performance of the TTS systems from multiple scenarios evaluated on
the ethnic language test data in L2 norm-squared error between the ground-truth and the predicted speech as a regression task. The
baseline model was supervisedly trained using only the speech-text paired data of the Indonesian language, and the performance reached
1.162 of the L2 norm-squared on average. For the proposed system, we observed similar trends with the ASR results, where
semi-supervised training with the speech chain method improved significantly over the baseline and achieved a L2 norm-squared
reduction to 0.537 of L2 norm-squared on average. This performance is close to the upper-bound result, which is 0.441 of the L2
norm-squared on average.

\section{Conclusion}
We introduce a cross-lingual machine speech chain approach to construct an ASR and a TTS for the following Indonesian ethnic
languages, Javanese, Sundanese, Balinese, and Bataks, when no paired speech or text data of those languages was available. We
first pre-trained the ASR and TTS systems on the standard Indonesian language with parallel speech-text in a supervised
manner. We then performed a speech chain mechanism with only limited text or limited speech of the Indonesian ethnic language
(unsupervised learning). Experimental results revealed that our proposed speech-chain model achieved better ASR and TTS
performances, indicating that such a closed-loop architecture enables ASR and TTS to teach each other and improved the
performance even without any paired data. Note that although this study only focuses on the cross-lingual approach of the Indonesian
language to Indonesian ethnic languages, the framework can be applied to any cross-lingual tasks without significant
modification. In the future, we will investigate with other indigenous languages.

\section{Acknowledgements}
Part of this work was supported by JSPS KAKENHI Grant Numbers JP17H06101
and JP17K00237.

\section{Bibliographical References}
\label{reference}
\bibliographystyle{lrec}
\bibliography{sltu_ethnic_spchain}

\section{Language Resource References}
\label{lr:ref}
\bibliographystylelanguageresource{lrec}
\bibliographylanguageresource{languageresource}

\end{document}